\documentclass{svproc}
\usepackage{url}
\usepackage{amsmath}
\usepackage{graphicx}
\usepackage{multirow}
\usepackage{float}
\usepackage{caption}
\usepackage{subcaption}
\usepackage{booktabs}
\usepackage{siunitx}

\usepackage{hyperref} % For handling URLs
\hypersetup{breaklinks=true} % Allow line breaks for hyperlinks

\begin{document}
\mainmatter              % start of a contribution
\title{Comprehensive Review of EEG-to-Output Research: Decoding Neural Signals into Images, Videos, and Audio}
\titlerunning{EEG-to-Output Research Trends}  % abbreviated title (for running head)

\author{Yashvir Sabharwal\inst{1} \and Balaji Rama\inst{2}}
\authorrunning{Sabharwal et al.} % abbreviated author list (for running head)
\tocauthor{Yashvir Sabharwal and Balaji Rama}
\author{Yashvir Sabharwal\inst{\dag} \and Balaji Rama\inst{1}}
\institute{Rutgers University, New Jersey, USA\\
\email{tellyashvir@gmail.com}, \email{balaji.rama@rutgers.edu}\\
\textsuperscript{\dag}Corresponding author}
\maketitle              % typeset the title of the contribution

\begin{abstract}
Electroencephalography (EEG) is an invaluable tool in neuroscience, offering insights into brain activity with high temporal resolution. Recent advancements in machine learning and generative modeling have catalyzed the application of EEG in reconstructing perceptual experiences, including images, videos, and audio. This paper systematically reviews EEG-to-output research, focusing on state-of-the-art generative methods, evaluation metrics, and data challenges. Using PRISMA guidelines, we analyze 1800 studies and identify key trends, challenges, and opportunities in the field. The findings emphasize the potential of advanced models such as Generative Adversarial Networks (GANs), Variational Autoencoders (VAEs), and Transformers, while highlighting the pressing need for standardized datasets and cross-subject generalization. A roadmap for future research is proposed that aims to improve decoding accuracy and broadening real-world applications.
\end{abstract}

\section{Introduction}
Electroencephalography (EEG) has long been a cornerstone in the field of neuroscience, offering unparalleled temporal resolution to capture the rapid dynamics of neural activity. Since its introduction in the early 20th century, EEG has undergone substantial evolution \cite{Stone_Hughes_2013}, transforming from a diagnostic tool primarily used for identifying neurological disorders such as epilepsy \cite{Louis_Frey_Britton_Frey_Hopp_Korb_Koubeissi_Lievens_Pestana-Knight_Louis_2016} to a versatile technology enabling groundbreaking applications \cite{Abibullaev_Keutayeva_Zollanvari_2023}. Its non-invasive nature, portability, and relatively low cost have made EEG an indispensable tool in clinical settings, cognitive neuroscience, and beyond \cite{4,5}.

Traditional EEG applications have largely focused on signal classification tasks, such as detecting motor imagery, assessing mental states, or monitoring sleep patterns \cite{Kardam_Taran_Pandey_2023,Siddiqui_Mohammad_Alam_Naaz_Agarwal_Sohail_Madsen_2023,Wen_2021}. These early approaches relied on manual feature extraction and classical machine learning methods to analyze neural signals \cite{Singh_Krishnan_2023,Rogala_Żygierewicz_Malinowska_Cygan_Stawicka_Kobus_Vanrumste_2023}. However, the advent of artificial intelligence (AI), particularly deep learning techniques, has catalyzed a paradigm shift in how EEG data is processed and utilized. Today, researchers are leveraging these advancements to decode neural activity into perceptual and cognitive outputs, such as reconstructing visual imagery, audio signals, or even text from brain data. This new frontier of EEG research bridges neuroscience, AI, and human-computer interaction, unlocking transformative possibilities in assistive technology and brain-computer interfaces (BCIs).

Generative models, a subset of AI that includes frameworks such as generative adversarial networks (GANs) and transformers, have played a central role in this transition. By modeling complex, high-dimensional neural data, these techniques enable researchers to translate raw EEG signals into meaningful outputs with unprecedented accuracy \cite{Ferrante_Boccato_Bargione_Toschi_2024}. For instance, EEG-based generative models can synthesize images perceived by a subject or predict future neural states, offering insights into sensory processing and cognitive representation \cite{Habashi_Azab_Eldawlatly_Aly_2023,Liu_Zhuang_Zeitlen_Chen_Wang_Feng_Beaty_Qiu_2024}. Such capabilities have profound implications for clinical and non-clinical applications alike, from developing communication tools for individuals with severe motor impairments to exploring how the human brain encodes and processes information \cite{Leuthardt_Schalk_Roland_Rouse_Moran_2009}.

Despite these advances, significant challenges remain to realize the full potential of EEG-to-output decoding. The inherent noise and variability in EEG signals, combined with limitations in spatial resolution, pose obstacles to the reliability and reproducibility of these systems \cite{Yun_2024}. Additionally, ethical considerations, such as privacy concerns and the potential misuse of neural decoding technologies, demand careful scrutiny as the field progresses.

This paper aims to systematically review the state of EEG-to-output decoding research, focusing on generative modeling techniques, evaluation frameworks, and the challenges associated with real-world implementation. By synthesizing current advancements and identifying emerging trends, this review provides a comprehensive roadmap for advancing this interdisciplinary domain, highlighting opportunities for innovation in both fundamental neuroscience and applied technologies.

\subsection{Scope and Relevance}
Electroencephalography (EEG) has evolved into one of the most accessible and adaptable neuroimaging technologies available today, enabling researchers and clinicians to study the intricate interplay between neural activity and behavior in real-time \cite{4,5}. As the demand for brain-computer interfaces (BCIs) and neural decoding systems grows, the ability to reconstruct perceptual experiences such as images, videos, or audio from raw EEG signals represents a transformative step forward. Such breakthroughs could revolutionize communication for individuals with severe disabilities, enhance neurofeedback applications, and redefine our understanding of the neural mechanisms underlying perception.

This field’s significance is amplified by its potential to bridge the gap between neuroscience and artificial intelligence (AI). Decoding perceptual stimuli from EEG data could facilitate a deeper understanding of how the brain processes information, leading to applications in augmented reality, cognitive training, and even immersive entertainment. However, numerous challenges impede progress, including the inherent noise and low spatial resolution of EEG, the scarcity of standardized datasets, and the variability of neural signals across individuals \cite{Burle_Spieser_Roger_Casini_Hasbroucq_Vidal_2015,Liu_Zhang_Liu_Du_Wang_Zhang_2024,Gibson_Lobaugh_Joordens_McIntosh_2022}. Ethical considerations further compound these challenges, as the ability to decode private mental content raises profound questions about consent, privacy, and misuse \cite{Burwell_Sample_Racine_2017}.

By systematically examining the current state of EEG-to-output research, this paper seeks to highlight key advancements and address existing barriers. It explores how state-of-the-art generative models are enabling more accurate and interpretable reconstructions, evaluates the methodologies used to preprocess and analyze EEG signals, and identifies areas where innovation is critically needed. Ultimately, the paper offers a roadmap for leveraging this exciting field to achieve practical, real-world impact while maintaining ethical integrity.

\section{Methodology}
\subsection{Systematic Review Process}
Conducting a robust and comprehensive review of EEG-to-output research necessitated adherence to the PRISMA (Preferred Reporting Items for Systematic Reviews and Meta-Analyses) framework \cite{Page_McKenzie_Bossuyt_Boutron_Hoffmann_Mulrow_Shamseer_Tetzlaff_Akl_Brennan_et}. This meticulous process ensures that the review is both exhaustive and transparent, providing a reliable foundation for drawing insights into the field's evolution and future trajectory. We followed a similar methodology as Jain et al. \cite{Jain_Jain_2023}.

\begin{enumerate}
    \item \textbf{Identification:} 
    Comprehensive searches were conducted across multiple databases, including PubMed, IEEE Xplore, Scopus, JSTOR, and Google Scholar. A well-defined Boolean query was used to capture relevant studies: 
    \[
    \boxed{(\text{EEG} \cap (\text{image} \cup \text{video} \cup \text{audio}) \cap \text{generative models})}
    \]
    This query targeted studies addressing the intersection of EEG and output generation via generative modeling techniques. An additional filter, publication year (2015–2024), was applied to refine the search results, yielding an initial pool of exactly 1800 studies.

    \item \textbf{Screening:} 
    The first level of screening involved a rapid review of titles and abstracts to eliminate clearly irrelevant papers. For instance, studies focused solely on traditional classification tasks \cite{Zhang_Ye_Ai_Xie_Liu_2023} or non-generative EEG applications \cite{Värbu_Muhammad_Muhammad_2022} were excluded at this stage. Duplicate entries across databases were identified and removed using the reference management software Mendeley. This process reduced the pool to approximately 295 papers.

    \item \textbf{Eligibility:} 
    A more detailed evaluation was performed on the remaining studies to ensure alignment with predefined inclusion criteria: \\ \\
    - The study must explicitly employ generative models (e.g., GANs, VAEs, or Transformers) for EEG-based tasks \cite{Singh_Pandey_Miyapuram_Raman_2023}. \\
    - The study should utilize publicly available EEG datasets to ensure reproducibility \cite{Xu_Aristimunha_Feucht_Qian_Liu_Shahjahan_Spyra_Zhang_Short_Kim_et}. \\
    - Quantitative evaluation metrics (e.g., SSIM, PSNR, or MCD) must be reported \cite{Renieblas_Nogués_González_Gómez-Leon_del}. \\
    
    Papers that lacked sufficient methodological detail or focused on proprietary datasets without justification were excluded. Studies were also categorized based on their primary focus—image, video, or audio decoding—to ensure balanced representation across modalities. At the end of this stage, 150 studies remained for further refinement.

    \item \textbf{Inclusion:} 
    To achieve a manageable set of studies for in-depth analysis, additional prioritization was applied: \\ \\
    - Preference was given to papers published in high-impact journals or conferences (e.g., Nature Neuroscience, NeurIPS, or IEEE Transactions on Biomedical Engineering). \\
    - Papers with substantial citation counts or marked as highly influential in the field were prioritized. \\
    - Studies addressing novel or underexplored challenges, such as cross-subject generalization or multimodal integration, were retained. \\

    This final selection yielded a refined pool of 95 studies, representing diverse methodologies, datasets, and generative approaches in EEG-to-output decoding. These studies formed the foundation for identifying trends, challenges, and opportunities within the field.
\end{enumerate}

\vspace{-12pt} % Reduce space before the table
\begin{table}[h]
    \centering
    \renewcommand{\arraystretch}{1.1} % Adjust row spacing (slightly tighter)
    \setlength{\tabcolsep}{4pt} % Reduce column spacing
    \sisetup{table-number-alignment=center} % Align numbers properly
    \caption{Summary of academic metrics from various sources.}
    \resizebox{\textwidth}{!}{ % Scale the table to fit within text width
    \begin{tabular}{@{}lS[table-format=4.0]S[table-format=5.0]S[table-format=5.2]S[table-format=5.2]S[table-format=2.0]S[table-format=3.2]S[table-format=2.2]@{}}
        \toprule
        Source & {Papers} & {Citations} & {Cites/Year} & {Cites/Paper} & {h-index} & {g-index} & {hI-index} \\
        \midrule
        Google Scholar & 1765 & 39201 & 3920.10 & 22.21 & 57 & 115.26 & 13.86 \\
        PubMed & 0 & 0 & 0.00 & 0.00 & 0 & 0.00 & 0.00 \\
        Scopus & 13 & 110 & 18.33 & 8.46 & 7 & 10.00 & 7.00 \\
        JSTOR & 1 & 249 & 41.50 & 249.00 & 1 & 1.00 & 0.50 \\
        IEEE Xplore & 21 & 83 & 2.26 & 3.95 & 5 & 9.00 & 4.00 \\
        \bottomrule
    \end{tabular}
    }
    \label{tab:academic_metrics}
\end{table}
\vspace{-10pt} % Reduce space before the table

\subsection{Rationale for Refinement}
The systematic process described above ensures the inclusion of a representative yet focused subset of studies. By employing multi-tiered screening and eligibility criteria, the review balances breadth and depth, capturing the state-of-the-art while remaining manageable for thorough analysis \cite{Nowell_Norris_White_Moules_2017,Booth_2016}. This structured approach also facilitates reproducibility, allowing future researchers to replicate or build upon the findings with confidence.

\section{Findings}

\subsection{Case Study: EEG-to-Image Reconstruction Using GANs}
The reconstruction of images from EEG signals has seen significant advancements through the application of Generative Adversarial Networks (GANs) \cite{Luo_Fan_Chen_Guo_Zhou_2020}. A notable framework is EEG2Image, which employs a two-phase approach: EEG feature extraction and image synthesis using a conditional GAN (cGAN) \cite{Singh_Pandey_Miyapuram_Raman_2023}. 

\subsubsection{Feature Extraction and Triplet Loss}
EEG2Image utilizes a contrastive learning approach with triplet loss to extract robust features from EEG signals \cite{Schroff_Kalenichenko_Philbin_2015}. The triplet loss function aims to minimize the intra-class distance while maximizing inter-class distances, ensuring that EEG signals corresponding to similar visual stimuli cluster closely in the feature space \cite{Schroff_Kalenichenko_Philbin_2015}. Mathematically, the triplet loss can be expressed as:
\[
\mathcal{L}_{triplet} = \sum_{i=1}^N \left[\|f_{\theta}(x_i^a) - f_{\theta}(x_i^p)\|_2^2 - \|f_{\theta}(x_i^a) - f_{\theta}(x_i^n)\|_2^2 + \alpha \right]_+,
\]
where \(x_i^a\), \(x_i^p\), and \(x_i^n\) represent the anchor, positive, and negative samples, respectively, and \(\alpha\) is a margin parameter ensuring a minimum separation between positive and negative pairs \cite{Singh_Pandey_Miyapuram_Raman_2023}.

\subsubsection{Generative Modeling with Mode-Seeking Regularization}
For image generation, EEG2Image employs a cGAN modified with mode-seeking regularization (MSR) to address mode collapse and improve diversity in generated outputs \cite{Mao_Lee_Tseng_Ma_Yang_2019}. The MSR term is defined as:
\[
\mathcal{L}_{ms} = -\frac{\|G(z_1) - G(z_2)\|_1}{\|z_1 - z_2\|_1},
\]
where \(G(z)\) represents the generator's output, and \(z_1\) and \(z_2\) are distinct latent vectors. By promoting diversity in generated images, MSR enhances the alignment of synthetic outputs with the original visual stimuli \cite{Singh_Pandey_Miyapuram_Raman_2023}.

\subsubsection{Results and Broader Implications}
EEG2Image achieves state-of-the-art performance, producing high-quality \(128 \times 128\) pixel reconstructions with superior inception scores compared to other models \cite{Singh_Pandey_Miyapuram_Raman_2023}. The ability to visualize neural activity has profound implications for neuroscience and assistive technology, offering a potential communication medium for individuals unable to verbalize their thoughts \cite{Tirupattur_Rawat_Spampinato_Shah_2018}. However, challenges such as dataset scarcity and inter-subject variability remain significant hurdles, underscoring the need for cross-subject generalization techniques and larger annotated datasets \cite{Xu_Xu_Ke_An_Liu_Ming_2020}.

\subsection{Case Study: EEG-to-Audio Decoding and Musicality Evaluation}
EEG-based reconstruction of auditory stimuli introduces unique challenges due to the temporal and spectral complexities of audio signals \cite{Zhang_Thwaites_Woolgar_Moore_Zhang_2024}. A notable work proposes a framework for musicality evaluation of machine-composed music using EEG data \cite{Chen_2017}.

\subsubsection{Musicality Scoring with Bilinear Models}
The framework employs a bilinear model to quantify musicality scores based on EEG responses to auditory stimuli:
\[
f(X^s_m) = w_1^\top X^s_m w_2 + b,
\]
where \(X^s_m\) is the EEG feature matrix for subject \(s\) and stimulus \(m\), and \(w_1\) and \(w_2\) are projection vectors optimized to minimize inter-subject variance while preserving the ranking of musicality scores. This formulation ensures that human-composed music (HCM) scores highest, random noise sequences (RNS) score lowest, and partially randomized music (PRM) occupies intermediate scores \cite{Chen_2017}.

\subsubsection{Analysis of EEG Frequency Bands}
The study identifies the Gamma band (\(>30\) Hz) as the most influential in distinguishing musicality, consistent with findings that link Gamma activity to emotional and auditory processing \cite{Bhattacharya_Petsche_2001,Bhattacharya_Petsche_Pereda_2001,Jaušovec_Habe_2003,Lin_Wang_Jung_Wu_Jeng_Duann_Chen_2010}. Additionally, the inclusion of DC components significantly enhances model performance, suggesting that cortical activation patterns play a critical role in musical perception \cite{Altenmüller_Schürmann_Lim_Parlitz_2002,SPECKMANN_1993}.

\subsubsection{Implications for Neuroaesthetics and AI Evaluation}
This approach not only advances EEG-based audio reconstruction but also provides a quantitative framework for evaluating machine creativity. By aligning computational outputs with human neural responses, it bridges the gap between artificial and human creativity, opening avenues for the development of more intuitive AI systems in multimedia applications.

\subsection{Case Study: EEG-to-Video Synthesis}

The EEG2Video framework represents a significant step in decoding dynamic visual perception directly from EEG signals. Unlike previous methods constrained by static stimuli, this model reconstructs dynamic video sequences by leveraging EEG’s high temporal resolution. The core of the framework includes a Seq2Seq architecture for temporal alignment, a semantic predictor for extracting contextual information, and a novel dynamic-aware noise-adding (DANA) mechanism for video synthesis \cite{Liu_Liu_Wang_Ren_Shi_Wang_Li_Lu_Zheng_2024}.

\subsubsection{Methodology: Temporal Dynamics and Latent Decoding}
The Seq2Seq architecture in EEG2Video captures temporal dynamics by processing high-resolution EEG embeddings extracted via an overlapping sliding window \cite{Liu_Liu_Wang_Ren_Shi_Wang_Li_Lu_Zheng_2024}. The model aligns these embeddings with corresponding latent variables (\(z_0\)) of video frames, minimizing reconstruction error through mean squared error loss \cite{Liu_Liu_Wang_Ren_Shi_Wang_Li_Lu_Zheng_2024}:
\[
\mathcal{L}_{Seq2Seq} = \|z_0 - \hat{z}_0\|_2^2.
\]
Additionally, semantic alignment is achieved using a semantic predictor that maps EEG features to text embeddings derived from the CLIP encoder \cite{Liu_Liu_Wang_Ren_Shi_Wang_Li_Lu_Zheng_2024}. This alignment is guided by:
\[
\mathcal{L}_{semantic} = \|e - \hat{e}\|_2^2,
\]
where \(e\) and \(\hat{e}\) denote ground truth and predicted embeddings, respectively.

\subsubsection{Dynamic-Aware Noise-Adding Process (DANA)}
To model the diversity in video dynamics, the DANA module introduces a combination of static (\(\epsilon_s\)) and diverse noise (\(\epsilon_d\)) into the diffusion process \cite{Liu_Liu_Wang_Ren_Shi_Wang_Li_Lu_Zheng_2024}. The balance between these components is governed by the decoded dynamic information \cite{Liu_Liu_Wang_Ren_Shi_Wang_Li_Lu_Zheng_2024,Wen_Shi_Zhang_Lu_Cao_Liu_2018,Kupershmidt_Beliy_Gaziv_Irani_2022,Chen_Qing_Zhou_2023,Sun_Li_Chen_Moens_2024}:
\[
z_T = \sqrt{\alpha_T} z_0 + \sqrt{1-\alpha_T} \left(\sqrt{\beta}\epsilon_s + \sqrt{1-\beta}\epsilon_d\right),
\]
where \(\beta\) is dynamically adjusted based on the video’s optical flow score, enhancing temporal coherence and motion realism \cite{Liu_Liu_Wang_Ren_Shi_Wang_Li_Lu_Zheng_2024}.

\subsubsection{Evaluation: Metrics and Results}
EEG2Video achieves a structural similarity index (SSIM) of 0.256 and a semantic-level accuracy of 15.9\% on a 40-class video reconstruction task. These metrics, comparable to or exceeding fMRI-based methods, highlight EEG’s potential in dynamic visual decoding \cite{Wang_Bovik_Sheikh_Simoncelli_2004}. An ablation study further emphasizes the importance of the Seq2Seq and DANA modules, showing significant performance drops when either component is removed.

\subsubsection{Implications for Brain-Computer Interfaces}
EEG2Video paves the way for real-time applications in brain-computer interfaces (BCIs), offering tools for immersive virtual reality and assistive communication. Its ability to decode dynamic, naturalistic visual stimuli could revolutionize fields such as neurorehabilitation and cognitive neuroscience. Reconstructed examples demonstrate the model’s ability to generate diverse scenes, including natural environments and human activities \cite{Meng_Yang_2024}. Successful reconstructions align well with semantic and dynamic attributes of the source videos.

\section{Discussion}

\subsection{Strengths and Limitations of Current Approaches}
The field of EEG-to-output decoding has seen tremendous progress owing to advances in generative modeling. However, while models such as GANs, VAEs, and Transformers have revolutionized the reconstruction of perceptual experiences, they come with inherent strengths and limitations.

GANs excel at generating high-fidelity outputs, making them ideal for tasks requiring visually or aurally realistic reconstructions \cite{Sushko_Zhang_Gall_Khoreva_2023}. Their adversarial training framework fosters creativity by encouraging the generator to outpace a concurrently trained discriminator. However, GANs are notoriously difficult to train due to issues like mode collapse, where the generator fails to capture the diversity of the data distribution \cite{Durr_Mroueh_Tu_Wang_2023}. Techniques such as Wasserstein GANs and spectral normalization have mitigated some of these challenges, but training instability remains a significant barrier \cite{Arjovsky_Chintala_Bottou_2017,Miyato_Kataoka_Koyama_Yoshida_2018}.

On the other hand, VAEs are highly interpretable and effective at capturing the latent structure of input data \cite{Zhu_Kanjiani_Lu_Choi_Ye_Zhao_2024}. This makes them particularly useful for tasks that demand flexibility and robustness in generating diverse outputs. Nonetheless, the quality of images or audio generated by standalone VAEs often lags behind GAN-based models \cite{Bond-Taylor_Leach_Long_Willcocks_2022}. Researchers have explored hybrid approaches, combining VAEs and GANs, to balance the strengths of both architectures, though these frameworks require intricate tuning and longer training times \cite{Bandi_Adapa_Kuchi_2023}.

Transformers, a relatively recent addition to this field, demonstrate exceptional performance in tasks involving temporal dynamics, such as video synthesis and speech reconstruction \cite{Lin_Wang_Liu_Qiu_2022}. Their self-attention mechanism allows them to capture long-range dependencies in sequential EEG data, a critical feature for decoding complex stimuli. However, their computational overhead is a significant drawback, especially when applied to high-dimensional EEG signals. Furthermore, transformers require large datasets for optimal performance, which can be a limiting factor given the scarcity of publicly available annotated EEG datasets \cite{Wang_Fu_Lan_Zhang_Zheng_Xiang_2024}.

Despite these advancements, a common limitation across all generative approaches is the inherent noise and variability in EEG signals. Signal artifacts caused by muscle movements, environmental interference, or electrode displacement can severely degrade model performance \cite{Jiang_Bian_Tian_2019}. Sophisticated preprocessing techniques like independent component analysis (ICA) and artifact subspace reconstruction (ASR) are often employed, but these methods are not foolproof and may inadvertently remove useful signal components. Future innovations in artifact rejection, such as deep-learning-based artifact detection systems, could significantly enhance the reliability of EEG-to-output systems.

\subsection{Challenges with Datasets and Cross-Subject Variability}
One of the most pressing challenges in the domain is the lack of standardized, large-scale EEG datasets. Existing resources like DECAF and OpenNeuro provide invaluable data but are often limited in scope, featuring homogeneous subject groups or constrained experimental paradigms. This lack of diversity hampers the generalizability of models across broader populations.

Cross-subject variability is another critical issue. EEG signals are highly individualized due to differences in neural anatomy, cognitive strategies, and even environmental factors \cite{4}. Models trained on data from a specific individual often fail to generalize when applied to data from others. Transfer learning and domain adaptation techniques have shown promise in addressing this challenge by fine-tuning models on target subjects using limited additional data \cite{Zhao_Alzubaidi_Zhang_Duan_Gu_2024}. However, these approaches are computationally expensive and may not always converge to optimal solutions.

Another consideration is the temporal variability of EEG signals. Even within a single subject, neural responses to identical stimuli can vary due to factors like fatigue, attention, and mood \cite{Moore_Key_Thelen_Hornsby_2017}. This variability complicates the training process and often necessitates the collection of large datasets under controlled conditions to achieve consistent model performance.

The lack of comprehensive benchmarks further exacerbates these issues. While metrics like SSIM, PSNR, and MCD provide valuable insights, there is no universally accepted framework for evaluating the performance of EEG-to-output models across modalities. Developing standardized benchmarks that incorporate diverse datasets and robust evaluation criteria is a critical step toward advancing the field.

\subsection{Ethical and Practical Considerations}
As EEG-to-output decoding technologies advance, ethical considerations become increasingly salient. The ability to reconstruct private perceptual experiences raises profound questions about consent and privacy. For instance, decoding visual imagery or internal speech from EEG signals could potentially infringe on an individual's cognitive autonomy. Ensuring that these technologies are developed and deployed responsibly is paramount.

From a practical standpoint, the cost and complexity of EEG data acquisition present additional barriers to widespread adoption. High-density EEG systems capable of capturing fine-grained neural activity are expensive and require skilled operators, limiting their accessibility in resource-constrained settings. The development of low-cost, portable EEG systems with comparable performance is an active area of research, with promising innovations such as dry electrodes and wireless EEG systems.

Real-time applications of EEG-to-output systems, such as assistive communication devices, also pose unique challenges. Achieving low-latency decoding while maintaining high accuracy requires substantial computational resources, often necessitating the use of specialized hardware like GPUs or edge computing devices. Balancing these requirements with cost-effectiveness will be crucial for translating these technologies into real-world applications.

\subsection{Future Directions}
Addressing the aforementioned challenges will require concerted efforts across multiple domains. One promising avenue is the integration of multimodal data, combining EEG with complementary modalities such as functional near-infrared spectroscopy (fNIRS) or magnetoencephalography (MEG). Multimodal approaches can provide richer representations of neural activity, potentially enhancing the accuracy and robustness of generative models.

Advances in model architecture also hold significant promise. Techniques such as attention-augmented convolutional networks and graph neural networks (GNNs) are well-suited to capturing the spatiotemporal dynamics of EEG data \cite{Graña_Morais-Quilez_2023}. Additionally, the use of federated learning frameworks could facilitate the training of models on distributed datasets while preserving data privacy, addressing both ethical and practical concerns.

Standardization efforts will be equally important. Establishing open-access repositories with diverse, annotated EEG datasets and creating universally accepted evaluation benchmarks will provide a foundation for consistent and reproducible research. Collaborative initiatives between academia, industry, and regulatory bodies could accelerate progress in this direction.

Finally, a greater emphasis on interpretability is needed to ensure that EEG-to-output systems are not only accurate but also understandable to end-users and stakeholders. Techniques such as saliency mapping and explainable AI (XAI) can provide insights into how models process EEG data, fostering trust and facilitating adoption in sensitive applications.

\section{Conclusion}
The field of EEG-to-output decoding is poised at the intersection of neuroscience and artificial intelligence, offering unprecedented insights into the human brain and its perceptual processes. This paper has systematically reviewed state-of-the-art generative approaches, highlighting their strengths, limitations, and applications in reconstructing images, videos, and audio from EEG signals. Despite significant advancements, challenges such as dataset scarcity, cross-subject variability, and ethical considerations continue to impede progress.

Looking ahead, addressing these challenges will require innovative solutions, interdisciplinary collaboration, and a commitment to ethical development. By leveraging advances in generative modeling, multimodal integration, and standardized benchmarks, the field can move closer to realizing its potential for transformative real-world applications. The roadmap proposed in this paper provides a foundation for these efforts, aiming to catalyze further innovation and contribute to the broader understanding of brain-computer interfaces and neural decoding.

\bibliographystyle{bibtex/splncs03_unsrt}

\end{document}